\documentclass[10pt,twocolumn,letterpaper]{article}

\usepackage{cvpr}
\usepackage{times}
\usepackage{epsfig}
\usepackage{graphicx}
\usepackage{amsmath}
\usepackage{amssymb}
\usepackage{multirow}

\usepackage[linesnumbered,ruled]{algorithm2e}
\usepackage{bm}
\usepackage{subfigure}

\usepackage[pagebackref=true,breaklinks=true,letterpaper=true,colorlinks,bookmarks=false]{hyperref}

\cvprfinalcopy 

\def\cvprPaperID{1077} 

\ifcvprfinal\fi
\begin{document}

\title{Deep Fitting Degree Scoring Network for Monocular 3D Object Detection}

\author{Lijie Liu$^{1,2,3,4}$, Jiwen Lu$^{1,2,3,*}$, Chunjing Xu$^{4}$, Qi Tian$^{4}$, Jie Zhou$^{1,2,3}$ \\
$^{1}$Department of Automation, Tsinghua University, China\\
$^{2}$State Key Lab of Intelligent Technologies and Systems, China\\
$^{3}$Beijing National Research Center for Information Science and Technology, China\\
$^{4}$Noah's Ark Lab, Huawei\\
{\tt\small liulj17@mails.tsinghua.edu.cn\qquad  \{lujiwen,jzhou\}@tsinghua.edu.cn,}\\
{\tt\small \{xuchunjing,tian.qi1\}@huawei.com}}

\maketitle
\thispagestyle{empty}
\newcommand\blfootnote[1]{%
\begingroup
\renewcommand\thefootnote{}\footnote{#1}%
\addtocounter{footnote}{-1}%
\endgroup
}
\blfootnote{$^{*}$   Corresponding Author}

\begin{abstract}
In this paper, we propose to learn a deep fitting degree scoring network for monocular 3D object detection, which aims to score fitting degree between proposals and object conclusively. Different from most existing monocular frameworks which use tight constraint to get 3D location, our approach achieves high-precision localization through measuring the visual fitting degree between the projected 3D proposals and the object. We first regress the dimension and orientation of the object using an anchor-based method so that a suitable 3D proposal can be constructed. We propose FQNet, which can infer the 3D IoU between the 3D proposals and the object solely based on 2D cues. Therefore, during the detection process, we sample a large number of candidates in the 3D space and project these 3D bounding boxes on 2D image individually. The best candidate can be picked out by simply exploring the spatial overlap between proposals and the object, in the form of the output 3D IoU score of FQNet. Experiments on the KITTI dataset demonstrate the effectiveness of our framework.
\end{abstract}

\section{Introduction}
2D perception is far from the requirements for people's daily use as people live in a 3D world essentially. In many applications such as autonomous driving \cite{bertozzi2000vision, chen2015deepdriving, janai2017computer, bai2018group, duan2018ai} and vision-based grasping \cite{saxena2008robotic, mahler2017dex, levine2018learning}, we usually need to reason about the 3D spatial overlap between objects in order to understand the realistic scene and take further action. 3D object detection is one of the most important problems in 3D perception, which requires solving a 9 Degree of Freedom (DoF) problem including dimension, orientation, and location. Although great progress has been made in stereo-based \cite{leibe2007dynamic, chen20153d, chen20183d}, RGBD-based \cite{lin2013holistic, song2014sliding, gupta2015aligning, song2016deep, qi2017frustum} and point-cloud-based \cite{wang2015voting, li2016vehicle, engelcke2017vote3deep, chen2017multi, li20173d, ku2017joint, asvadi2017depthcn, beltran2018birdnet, zeng2018rt3d} 3D object detection methods, monocular-image-based approaches have not been thoroughly studied yet, and most of existing works focus on the sub-problem, such as orientation estimation \cite{chen2016monocular, xiang2017subcategory, mousavian20173d}. The primary cause is that under monocular setting, the only cue is the appearance information in the 2D image, and the real 3D information is not available, which makes the problem ill-conditioned. However, in many cases, such as web images, mobile applications \cite{duan2016overview}, and gastroscopy, the information of depth or point-cloud is not available or unaffordable. Moreover, in some extreme scenarios, other sensors can be broken. Therefore, considering the rich source of monocular images and the robustness requirements of the system, monocular 3D object detection problem is of crucial importance.

\begin{figure}[tb]
    \begin{center}
       \includegraphics[width=1.0\linewidth]{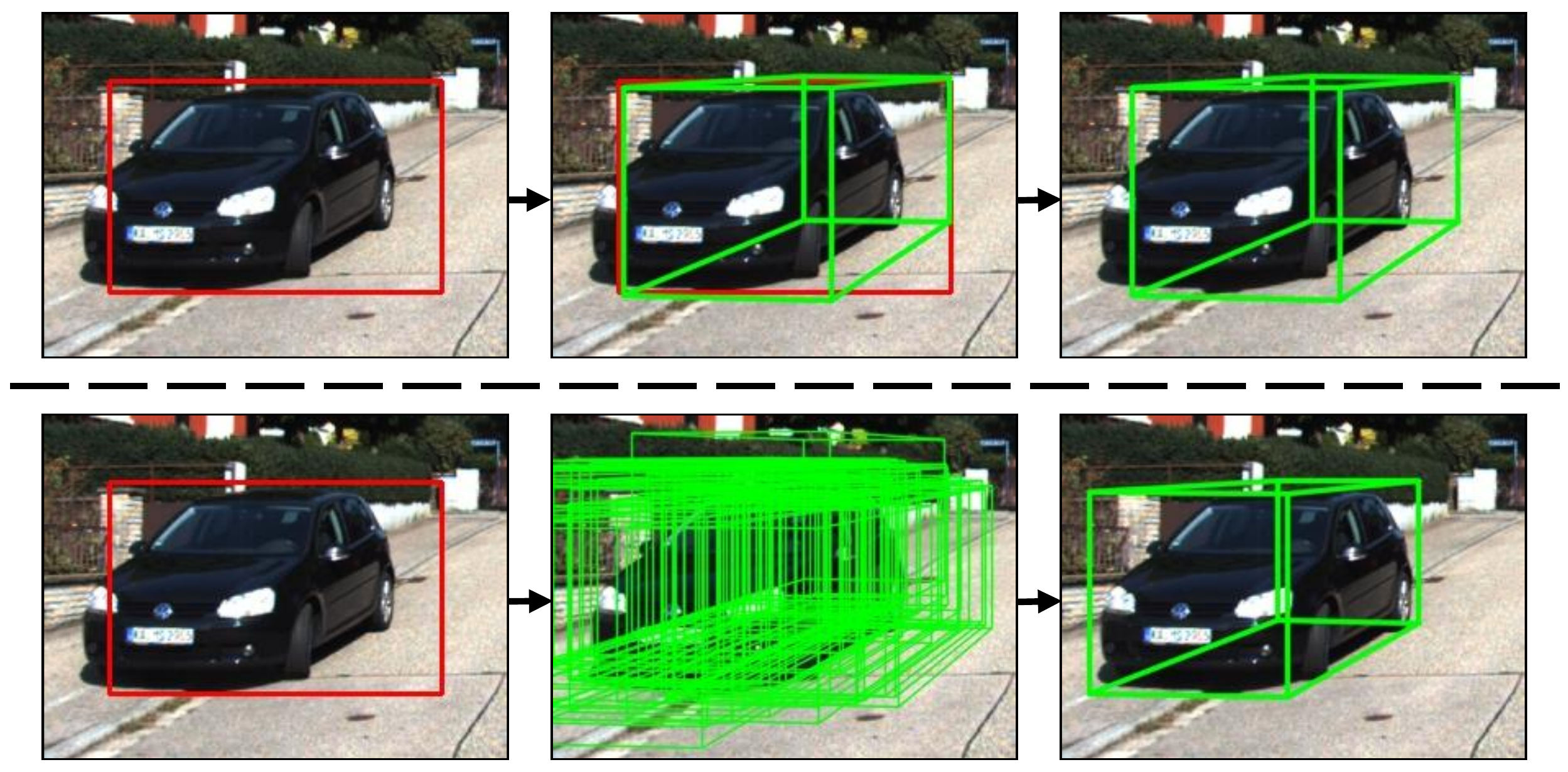}
    \end{center}
       \caption{Comparison between our proposed method and tight-constraint-based method. The upper part is the commonly used approach by many existing methods, which neglects the spatial relation between 3D projection and object, and is very sensitive to the error brought by 2D detection. The lower part is our proposed pipeline which reasons about the 3D spatial overlap between 3D proposals and object so that it can get better detection result.}
    \label{fig:idea}
\end{figure}

In monocular 3D object detection problem, dimension and orientation estimation are easier than location estimation, because the only available information, appearance, is strongly related to the former two sub-problems. On the contrary, it is not practical to directly regress location using a single image patch, since nearby and faraway objects with same pose are substantially identical in appearance.

Tight constraint~\cite{mousavian20173d, kundu20183d} is a commonly used method in monocular 3D object detection problem, which solves the location by placing the 3D proposal in the 2D bounding box compactly. However, the tight constraint has two significant drawbacks: 1) Image appearance cue is not used; thus it cannot benefit from a large number of labeled data in the training set. 2) Its performance highly depends on the 2D detection accuracy, as shown in Figure \ref{fig:idea}.

Inspired by the observation that people can easily distinguish the quality of 3D detection results through projecting these 3D bounding boxes on the 2D image and checking the relation between projections and object (fitting degree), we believe that exploring the 3D spatial overlap between proposals and ground-truth is the key to solve the location estimation problem. In this paper, we first regress the dimension and orientation of the object using an anchor-based method so that we can construct a suitable 3D proposal. The reason why we emphasize the importance of the regression step is that without an appropriate proposal, checking the fitting degree is impractical. Then, we propose Fitting Quality Network (FQNet) to infer 3D Intersection over Union (IoU) between 3D proposals and object only using 2D information. Our motivation is that though the 3D location is independent of 2D appearance, drawing the projection results on the 2D image can bring additional information for the convolutional neural network (CNN) to better understand the spatial relationship between the original 3D bounding boxes and the object. As long as the network learns the pattern of the projected 3D bounding boxes, it can gain the power of judging the relation between 3D proposals and the object, and achieve high-precision 3D perception. Figure~\ref{fig:idea} gives an illustration of the essential difference between our idea and existing tight constraint-based method. We can see that our method is not sensitive to the error of 2D detection results. To the best of our knowledge, we are the first to solve the monocular 3D detection problem by exploring the fitting degree between 3D proposals and object. We conducted experiments on the challenging KITTI dataset and achieved state-of-the-art monocular 3D object detection performance, which demonstrates the effectiveness of our framework.

\section{Related Work}

\noindent
\textbf{Monocular 3D Object Detection: }Monocular 3D object detection is much more difficult than 2D object detection because of the ambiguities arising from 2D-3D mapping. Many methods have taken the first step, which can roughly categorize into two classes: handcrafted approaches and deep learning based approaches.

Most of the early works belong to the handcrafted approaches, which concentrated on designing efficient handcrafted features. Payet and Todorovic~\cite{payet2011contours} used image contours as basic features and proposed mid-level features, called bags of boundaries (BOBs). Fidler~\emph{et al.}~\cite{fidler20123d} extended the Deformable Part Model (DPM) and represented an object class as a deformable 3D cuboid composed of faces and parts. Pepik~\emph{et al.}~\cite{pepik2015multi} included viewpoint information and part-level 3D geometry information in the DPM and achieved robust 3D object representation. Although these handcrafted methods are very carefully designed and perform well on some scenarios, their generalization ability is still limited.

\begin{figure*}[htb]
    \begin{center}
       \includegraphics[width=1.0\linewidth]{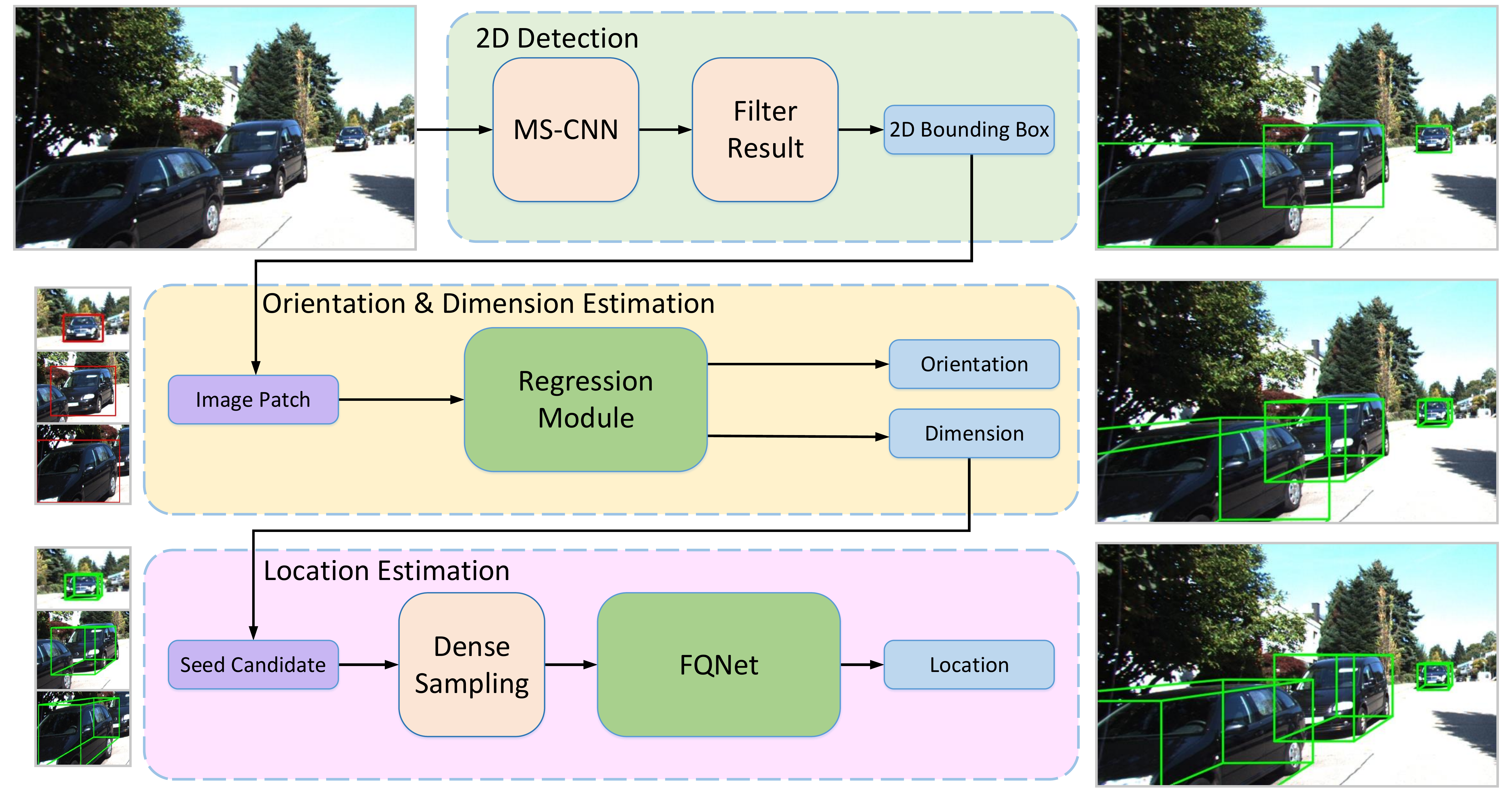}
    \end{center}
       \caption{The overall pipeline of our proposed monocular 3D object detection method, which only requires a single RGB image as input, and can achieve a 3D perception of the objects in the scene. We show some intermediate results on the right.
}
    \label{fig:framework}
\end{figure*}

Deep learning based approaches aim to benefit from end-to-end training and a large amount of labeled data. Chen~\emph{et al.}~\cite{chen2016monocular} generated a set of candidate class-specific object proposals on a ground prior and used a standard CNN pipeline to obtain high-quality object detections. Mousavian~\emph{et al.}~\cite{mousavian20173d} presented MultiBin architecture for orientation regression and tight constraint to solve the 3D translation. Kundu~\emph{et al.}~\cite{kundu20183d} trained a deep CNN to map image regions to the full 3D shape and pose of all object instances in the image. Apart from these pure monocular methods, there are some other methods which use additional information for training. Xu and Chen~\cite{xu2018multi} proposed to fuse a monocular depth estimation module and achieved high-precision localization. Chabot~\emph{et al.}~\cite{chabot2017deep} presented Deep MANTA (Deep Many-Tasks) for simultaneous vehicle detection, part localization and visibility characterization, but their method requires part locations and visibility annotations. 
In this paper, we propose a unified deep learning based pipeline, which does not require additional labels and can be trained end-to-end using a large number of augmented data.
\newline\newline
\noindent
\textbf{Box Refinement Techniques: }Our work has some similarities to the box refinement techniques, which focused on improving the localization accuracy. In 2D object detection, the most common method is the bounding box regression, which was first proposed by Felzenszwalb~\emph{et al.}~\cite{felzenszwalb2010object} and has been used in many state-of-the-art detectors, such as Faster R-CNN \cite{ren2015faster} and SPP-net \cite{he2015spatial}. Gidaris and Komodakis~\cite{gidaris2016locnet} proposed LocNet to further improve the object-specific localization accuracy through assigning probabilities to the boundaries. While in monocular 3D object detection, the work on this level has been limited. Xiao~\emph{et al.}~\cite{xiao2012localizing} proposed to localize the corners using a discriminative parts-based detector. Many works also used stronger representations to achieve high-precision localization. For example, Zia~\emph{et al.}~\cite{zia2015towards} used a fine-grained 3D shape model, Xiang and Savarese \cite{xiang2013object} introduced 3D aspectlet based on a piecewise planar object representation and Pero~\emph{et al.}~\cite{del2013understanding} proposed to use detailed geometric models. In our case, we stick to the 3D bounding box representation and learn the pattern of projected boxes on the 2D image.

\section{Approach}
Our framework only requires a single image as input and can output precise 3D detection results including dimension, orientation, and location of interested objects. Figure~\ref{fig:framework} shows the overall pipeline, where we first perform regular 2D detection\footnote{We use a popular 2D detection algorithm \cite{cai2016unified} to produce the 2D detection results.} and then use an anchor-based regression module to regress the dimension and orientation of each object based on the image patch cropped by 2D detection results. For the 3D location, we first use tight constraint to get a seed-candidate, then perform Gaussian dense sampling to generate a large number of candidates within a small range around the seed-candidate. To evaluate these candidates, we train a Fitting Quality Network (FQNet) to infer the 3D IoU between a large number of augmented samples and the ground truth. Therefore, by estimating the fitting degree between the candidates and the object, the candidate with the highest score will be chosen as the 3D detection result. Our framework separates the dimension and orientation estimation process from the location estimation because we consider that these tasks are fundamentally different (orientation and dimension are all appearance-related but the location is not) and the results of dimension and orientation regression have a significant impact on the location estimation process.

\subsection{Regression Module}
The input of our regression module is the cropped detection result of 2D object detection, while the output is the dimension and orientation of each object.
For dimension, based on appearance, it is not difficult to infer the type of cars, and cars of the same type usually have similar length, width, and height. For orientation, it is a bit more complicated since there are global orientation and local orientation, but intuitively we can be sure that different orientation will show different appearance. 
\newline\newline
\noindent
\textbf{Dimension Estimation: }To regress the dimension as accurate as possible, we propose an idea called anchor cuboid, whose philosophy is similar to MultiBin \cite{mousavian20173d}. We first perform k-means clustering on the training dataset to find the $K$ cluster centers of the dimension and regard these cluster centers as 3D anchor cuboids. During the regression process, the regression module outputs confidence and offset for each 3D anchor cuboid respectively, so the output is a $4K$-dimensional vector ($[c_i,\Delta w_i,\Delta h_i,\Delta l_i],i=1,...,K$), and the final regression result is the anchor cuboid with the highest confidence adding corresponding offsets. We optimize this module using the following loss function:
\begin{equation}
    L_d = - \log \sigma(c_{i^\star})+
		[ 1 - IoU(\bm{A}_{i^\star} + [\Delta w_{i^\star},\Delta h_{i^\star},\Delta l_{i^\star}] , \bm{G}) ]
\label{eqa:ld}
\end{equation}
where the ${i^\star}$ \footnote{It is worth mentioning that the value of ${i^\star}$ for each object can be computed and saved in the dataset before the training process.} means among $K$ anchor cuboid, the ${i^\star}$-th anchor cuboid $\bm{A}_{i^\star}$ has the maximum IoU with the ground-truth cuboid $\bm{G}$, and $[\Delta w_{i^\star},\Delta h_{i^\star},\Delta l_{i^\star}]$ are the offsets in three different dimensions relative to anchor cuboid $\bm{A}_{i^\star}$.
The $\sigma(\cdot)$ is the softmax function:
\begin{equation}
    \sigma(c_{i^\star}) = \frac{e^{c_{i^\star}}}{\sum_{i=1}^K e^{c_i}}
\end{equation}
and the function $IoU(\cdot,\cdot)$ computes the 3D IoU between two center-aligned cuboids:
\begin{equation}
    IoU(\bm{A},\bm{B}) = \frac{volume(\bm{A}) \cap volume(\bm{B})}{volume(\bm{A}) \cup volume(\bm{B})}
\end{equation}

\begin{figure}[htbp]
\centering
\subfigure[]{
\begin{minipage}{0.50\linewidth}
\centering
\includegraphics[width=1.0\linewidth]{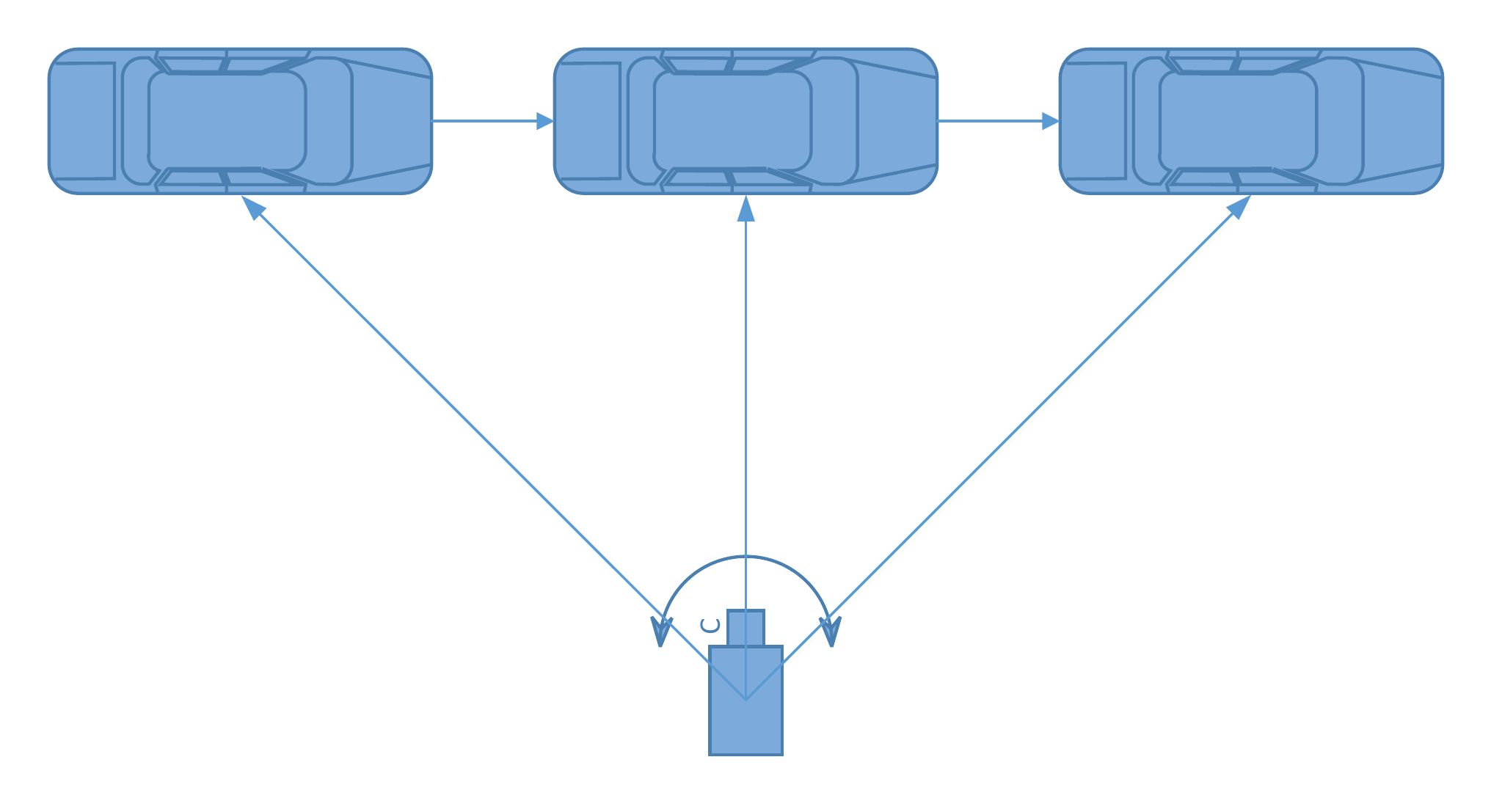}
\end{minipage}%
}
\subfigure[]{
\begin{minipage}{0.44\linewidth}
\centering
\includegraphics[width=1.0\linewidth]{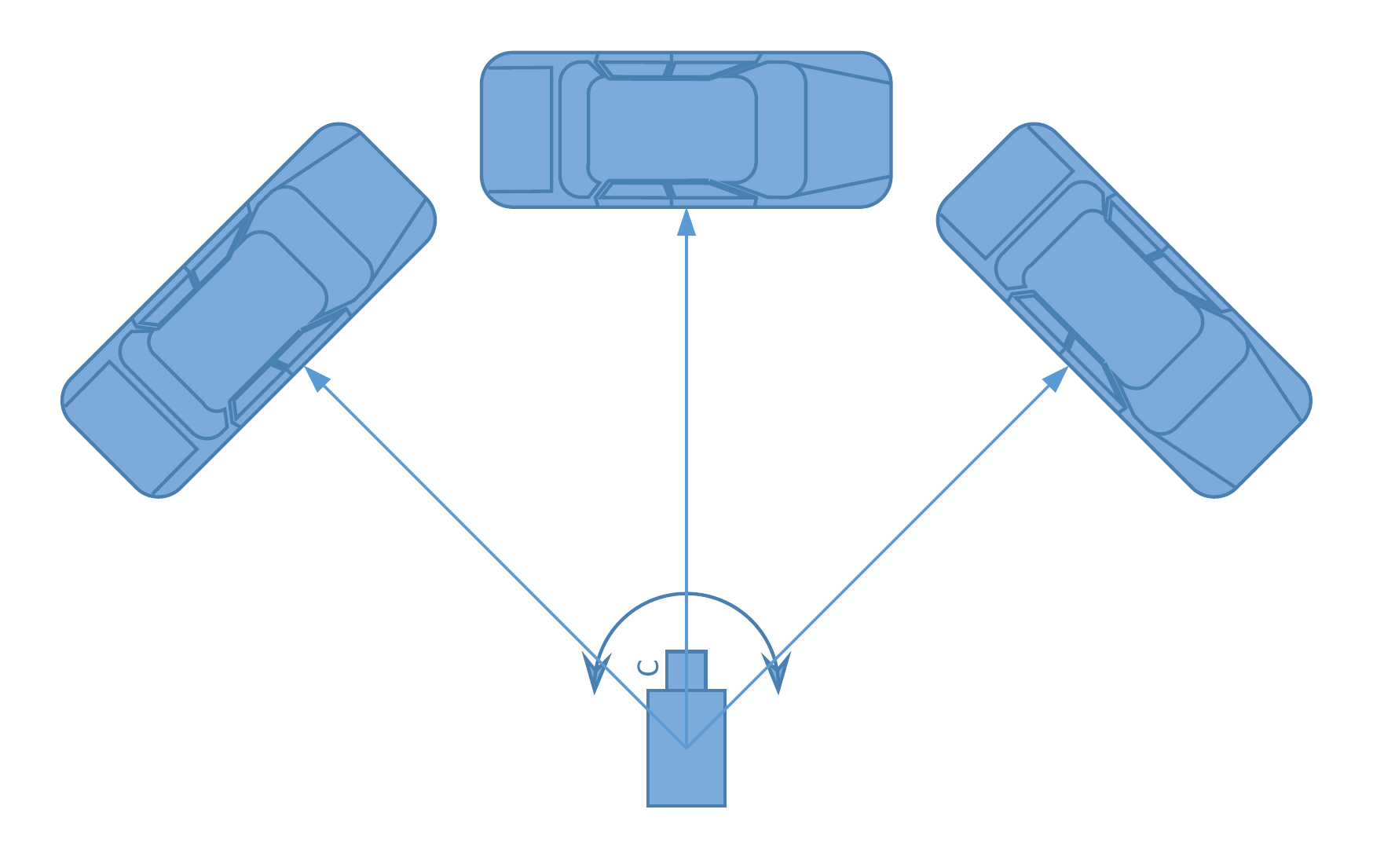}
\end{minipage}
}
\caption{In (a), the global orientations of the car are all facing the right, but the local orientation and appearance will change when the car moves from the left to the right. In (b), the global orientations of the car differ, but both the local orientation in the camera coordinates and the appearance remain unchanged. Hence, we can see that the appearance only has a relationship with the local orientation, and we can only regress the local orientation of the car based on the appearance. If we want to compute the global orientation using local orientation, we need to know the ray direction between the camera and the object, which can be calculated using the location of the object in the 2D image, more details are included in the supplementary materials.}
\label{fig:orien}
\end{figure}

There are two terms in the Equation (\ref{eqa:ld}). For the first term, it encourages the module to give the highest confidence to the anchor cuboid which has the maximum IoU with the ground truth dimension $G$, and provides low confidence for other anchor cuboids at the same time. For the second term, it encourages the module to regress the offset between the best anchor cuboid and ground truth cuboid. Our loss function is volume-driven rather than dimension-driven, and the rationale is that it can take into account the information from three dimensions synthetically and avoid the situation that two dimensions have good estimation while one dimension is not well estimated.
\newline\newline
\noindent
\textbf{Orientation Estimation: }There are two orientations in the 3D object detection problem, global orientation and local orientation. The global orientation of an object is defined under the world coordinates and will not change with the camera pose, while the local orientation is defined under the camera coordinates and hinges on how the camera shots the object. Figure \ref{fig:orien} gives an illustration of these two kinds of orientation from the bird's view. In this paper, we focus on the estimation of the local orientation, which is evaluated in the KITTI dataset and directly related to the appearance. 

\begin{figure}[htb]
    \begin{center}
       \includegraphics[width=1.0\linewidth]{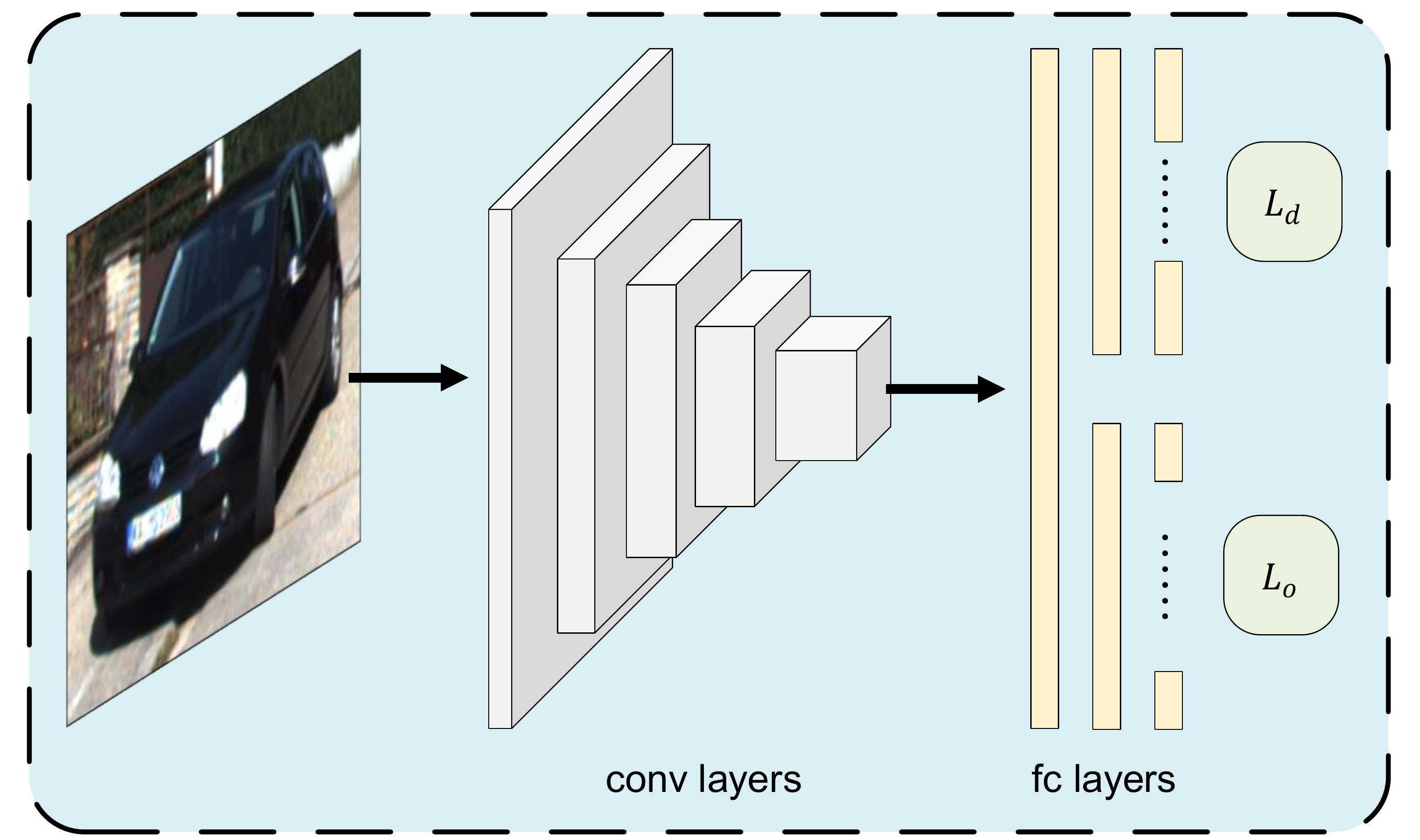}
    \end{center}
       \caption{The architecture of our regression module. There are two branches in the fully connected layers, for dimension regression and orientation regression respectively.}
    \label{fig:regression}
\end{figure}

\begin{figure*}[tb]
    \begin{center}
       \includegraphics[width=1.0\linewidth]{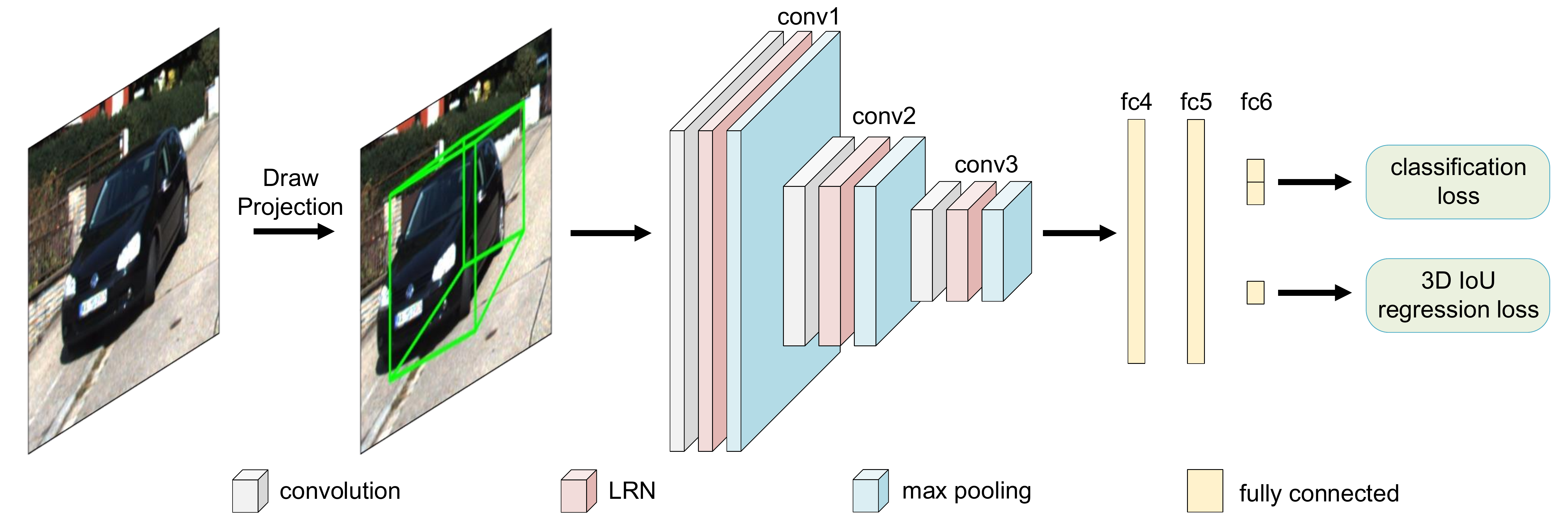}
    \end{center}
       \caption{The detailed architecture of FQNet. During pre-training process, the classification loss is used to train the convolutional layers and fully connected layers. The core innovation of our proposed FQNet is the input part, where our input image has additional artificial information.}
    \label{fig:FQNet}
\end{figure*}

For orientation regression, the range of the output is $[-\pi, \pi]$, we use a similar idea to dimension regression and perform k-means clustering on the training set to get $K'$ cluster centers. The output is a $2K'$-dimensional vector ($[c'_i,\Delta \theta_i],i=1,...,K'$). We define the loss function as follows:
\begin{equation}
    L_o = - \log \sigma(c_{i^\star})+
		[1-\cos(\Theta_{i^\star} +\Delta \theta_{i^\star} - \theta_G)]
\end{equation}
where $\Theta_{i^\star}$ is the nearest anchor angle comparing to the ground truth local orientation  $\theta_G$. The first term in the loss function $L_o$ is the same as $L_d$ which encourages the module to give high confidence to the nearest anchor angle, and the second term uses the cosine function to ensure that the offset $\theta_{i^\star}$ can be well regressed.

The idea behind our anchor-based regression is that directly regress a continuous variable is very difficult because the value range is  vast. Using an anchor-based method, we can first solve a classification problem to choose the best anchor, and then regress an offset based on the anchor value; hence the value range that we need to regress can be significantly reduced. We show the concrete architecture of our proposed regression module in Figure \ref{fig:regression}.

\subsection{Location Estimation}
After estimating the dimension and orientation, we can construct a 3D cuboid in 3D spatial space. We set the original coordinate of 8 vertices of the cuboid as:
\begin{equation}
\left\{
\begin{aligned}
    	\bm{x} &= \begin{bmatrix}\frac{l}{2}&\frac{l}{2}& -\frac{l}{2}& -\frac{l}{2}&\frac{l}{2}& \frac{l}{2}& -\frac{l}{2}& -\frac{l}{2}\end{bmatrix}\\
	\bm{y} &= \begin{bmatrix}0&0&0&0&-h&-h&-h&-h\end{bmatrix}\\
	\bm{z} &= \begin{bmatrix}\frac{w}{2}& -\frac{w}{2}& -\frac{w}{2}& \frac{w}{2}&\frac{w}{2}& -\frac{w}{2}& -\frac{w}{2}& \frac{w}{2}\end{bmatrix}
\end{aligned}
\right.
\end{equation}
where $l,h,w$ are the dimensions of the object. Assume that the 3D location is $\bm{T} = [T_x,T_y,T_z]^T$ in the camera coordinate, according to the law of camera projection, we have  
\begin{equation}
    \bm{K} \begin{bmatrix}
 \bm{R} & \bm{T} \\
 \bm{0}^T & 1
 \end{bmatrix} \begin{bmatrix}
   x_i  \\
   y_i  \\
   z_i  \\
   1
  \end{bmatrix}   = s \begin{bmatrix}u_i\\ v_i\\ 1\end{bmatrix}
\end{equation}
where $u_i$ and $v_i$ are the 2D projected coordinates of the $i$-th vertices, $\bm{K}$ is the intrinsic matrix, and $\bm{R}$ is the rotation matrix given by the global orientation $\theta$:
\begin{equation}\bm{R} = \begin{bmatrix}
 \cos \theta & 0& \sin \theta \\
0& 1& 0\\
-\sin \theta & 0& \cos \theta
 \end{bmatrix} 
\end{equation}
We can use OpenCV toolkit to draw the projected 3D bounding box on the 2D image based on the 2D projected coordinates; thus, our FQNet can learn from these 2D projection patterns and obtain the ability to reason 3D spatial relations.  
\newline\newline
\noindent
\textbf{Dense Sampling: }
We use a sampling and evaluation framework. Sampling in the whole 3D space is time-consuming, so we choose to get an approximate location (seed candidate) first. Many methods can achieve this goal, such as performing uniform sampling and searching the proposal whose 2D projection has the largest overlap with the 2D detection result. Here, we choose to use tight constraint similar in~\cite{mousavian20173d, kundu20183d} to locate the seed candidate, because it is fast and relatively more accurate. After computing the location of seed-candidate, we can perform dense sampling within a small range around it. We model the distribution of the transition value in three axes as three independent Gaussian distribution as follows:
\begin{equation}
\left\{
\begin{aligned}
    	\Delta{x} &\sim N(\mu_x,\sigma_x)\\
	\Delta{y} &\sim N(\mu_y,\sigma_y)\\
	\Delta{z} &\sim N(\mu_z,\sigma_z)
\end{aligned}
\right.
\end{equation}
where the mean and variance are all estimated using the 3D localization error in the training set. Thus, the $i$-th generated sample can be represented as $\bm{S}_i(x + \Delta{x_i},y + \Delta{y_i},z + \Delta{z_i}, l, h, w, \theta)$.
\newline\newline
\noindent
\textbf{FQNet: }The goal of FQNet is to evaluate the fitting quality between each sample $\bm{S}_i$ and object. Even though it is  challenging for CNN to infer the spatial relation between $\bm{S}_i$ and ground-truth 3D location of the object, after projecting the sample $\bm{S}_i$ on the 2D image, CNN can obtain this ability through learning the pattern that how well the edges and corners of projections are aligned with the specific part of the object. For quantitative analysis, we force the network to regress the 3D IoU between samples and the object. 
Denote the object image patch as $\bm{I}$, the objective of FQNet can be formulated as follows:
\begin{equation}
    \Theta^\star = \arg\min\limits_{\Theta} || \mathcal{F}(\bm{I},\bm{S}_i|\Theta) - IoU(\bm{I},\bm{S}_i)||
\end{equation}
where $\Theta$ denotes the parameters of FQNet.

Since we have the ground-truth 3D location of each object in our training set, we can generate an almost unlimited number of samples by adding a known jitter to the original 3D location. One problem is that how can we guarantee that FQNet can capture the pattern of the projection, which we painted on the 2D image manually. To answer this question, we first pre-train our FQNet to execute a classification task, in which it has to decide whether the input image patch contains an artificially painted 3D bounding box projection or not. The architecture of our proposed FQNet and its pre-training version are shown in Figure~\ref{fig:FQNet}. For the pre-training process, we use the cross-entropy loss for the classification task. For the 3D IoU regression, we use the smooth L1 loss, since it is less sensitive to outliers compared to the L2 loss.

\begin{figure}[tb]
    \begin{center}
       \includegraphics[width=1.0\linewidth]{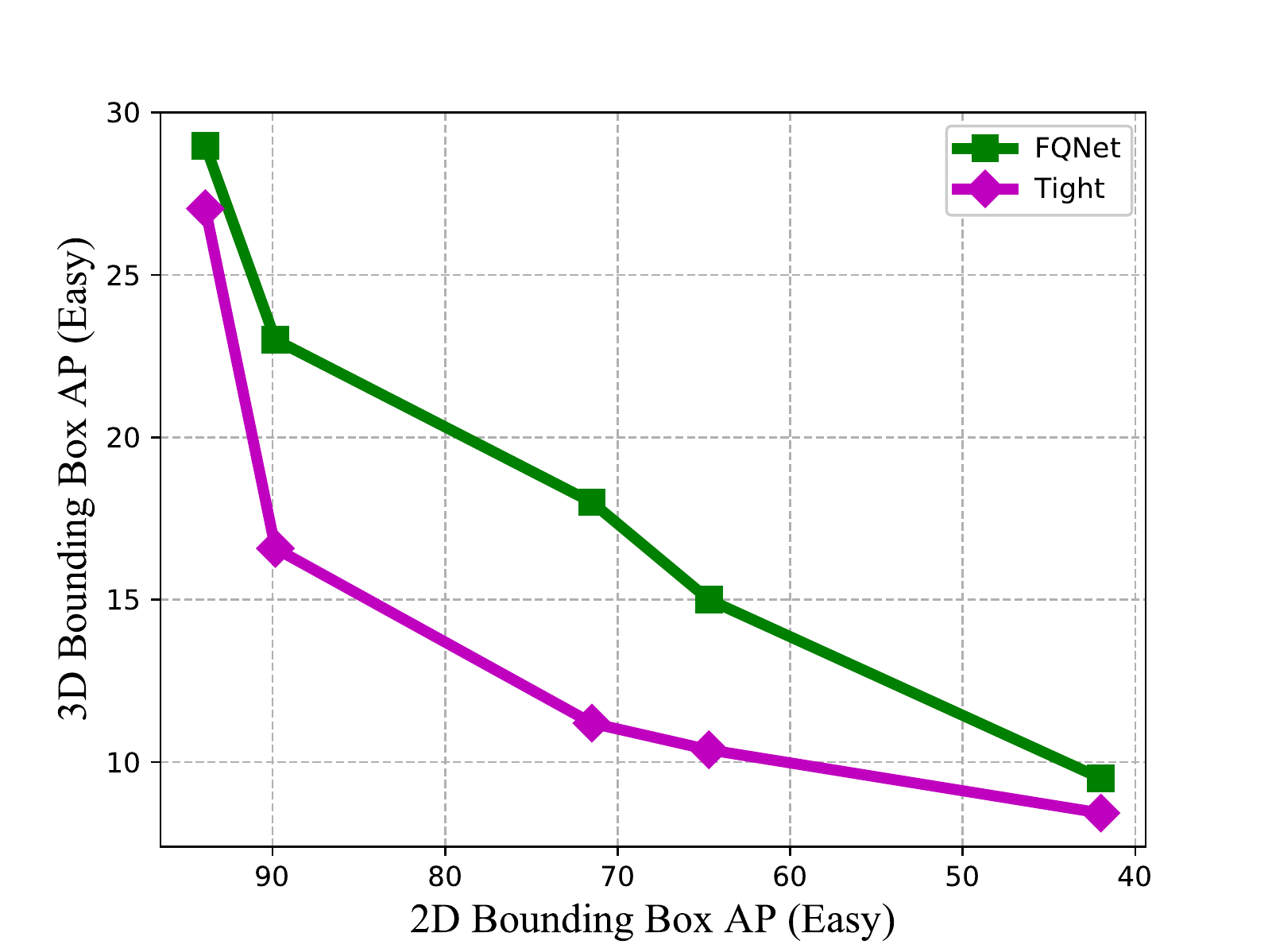}
    \end{center}
       \caption{3D detection performance comparison between our  with the tight-constriant-based baseline.}
    \label{fig:sensitive}
\end{figure}

\section{Experiments}
We evaluated our method on the real-world KITTI dataset \cite{geiger2012we}. The KITTI object detection benchmark includes 2D Object Detection Evaluation, 3D Object Detection Evaluation and Bird's Eye View Evaluation. There are 7481 training images and 7518 testing images in the dataset, and in each image, the object is annotated with observation angle (local orientation), 2D location, dimension, 3D location, and global orientation. However, only the labels in the KITTI training set are released, so we mainly conducted controlled experiments in the training set. Results are evaluated based on three levels of difficulty: Easy, Moderate, and Hard, which is defined according to the minimum bounding box height, occlusion and truncation grade. There are two commonly used train/val experimental settings: Chen~\emph{et al.}~\cite{chen20153d, chen2016monocular} (train/val 1) and Xiang~\emph{et al.}~\cite{xiang2015data, xiang2017subcategory} (train/val 2). Both splits guarantee that images from the training set and validation set are from different videos. We focused our experiment on the Car object category since, for Pedestrian and Cyclist, there is not enough data to train our model. 

\begin{figure}[tb]
    \begin{center}
       \includegraphics[width=1.0\linewidth]{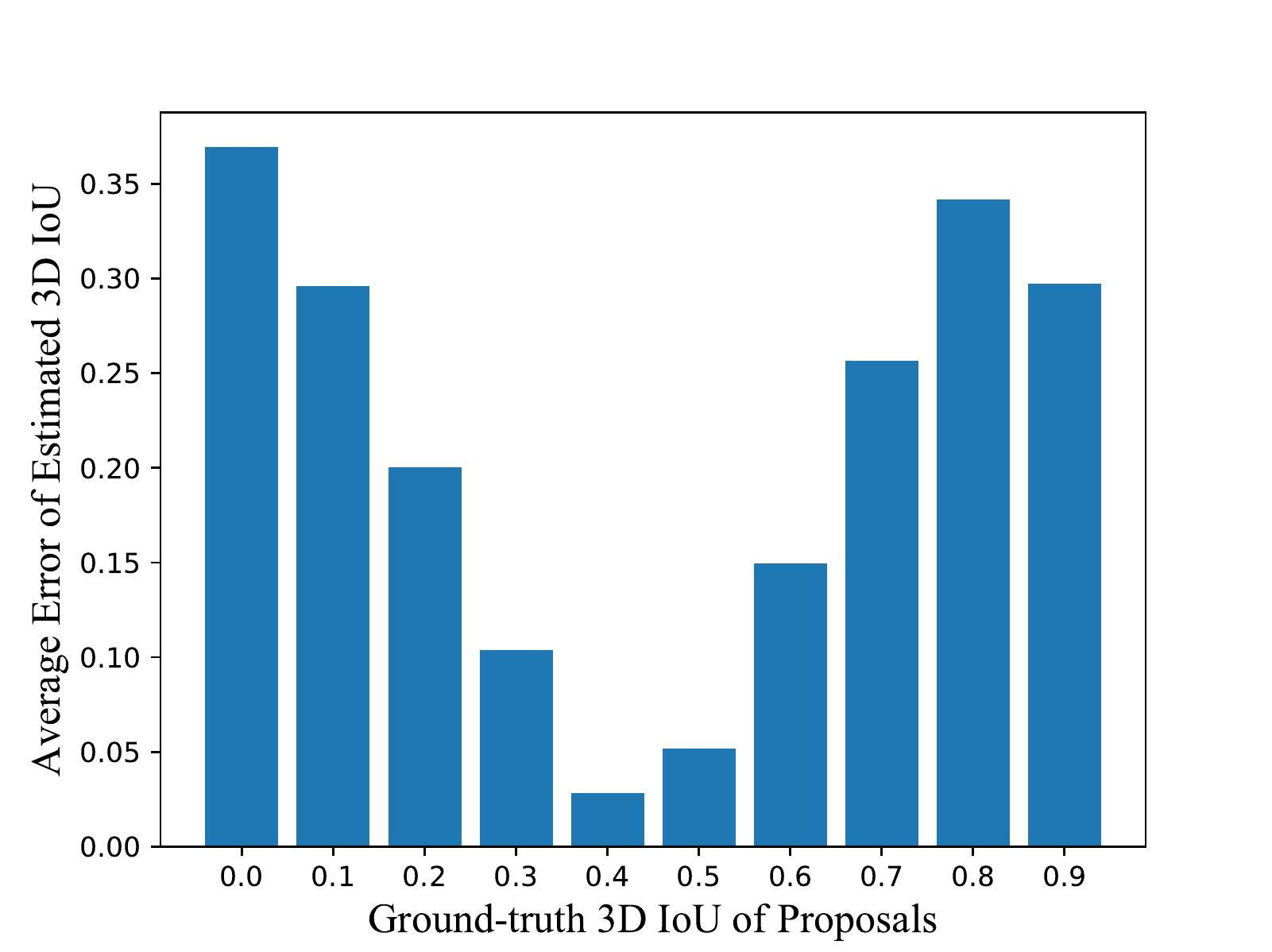}
    \end{center}
       \caption{Histogram of average regression error for proposals from ten levels of 3D IoU.}
    \label{fig:iou}
\end{figure}

\subsection{Implementation Details}

For regression module, we used the ImageNet \cite{deng2009imagenet} pre-trained VGG-16 \cite{simonyan2014very} model with $224\times 224$ input size to initialize the weights of convolutional layers. 
We used four anchor cuboids during dimension estimation and two anchor angles during orientation estimation. The module is trained with SGD using a fixed learning rate of $10^{-4}$ with a batch size of 8. We performed data augmentation by adding color distortions, flipping  images at random, and jittering 2D boxes with a translation of $0\sim 0.03$x height and width. 

For dense sampling process, 
we first sampled 1024 samples around the seed candidate. After discarding the samples more than half of which are outside the image plane, we kept 640 samples for evaluation.

\begin{table*}[tb]
\caption{Comparisons of the Average Orientation Similarity (AOS) with the state-of-the-art methods on the KITTI dataset.}
\vspace{-0.35cm}
\begin{center}
\begin{tabular}{|c|c|c|c|c|c|c|c|c|c|}
\hline
\multirow{2}*{Method} &  \multicolumn{3}{c|}{Easy} & \multicolumn{3}{c|}{Moderate} & \multicolumn{3}{c|}{Hard}\\
\cline{2-10}
~ &  train/val 1 & train/val 2& test & train/val 1 & train/val 2& test   &train/val 1 & train/val 2& test   \\
\hline\hline
3DOP \cite{chen20153d} &91.58 & -&91.44 & 85.80 & -&86.10 & 76.80  & -&76.52\\
Mono3D \cite{chen2016monocular} &91.90 & - &91.01& 86.28 & -&86.62 & 77.09  & -&76.84\\
3DVP \cite{xiang2015data} &- & 78.99 &86.92&- & 65.73&74.59 & -  & 54.67&64.11\\
SubCNN \cite{xiang2017subcategory} &- & 94.55&90.67 &- & 85.03&88.62 &-  & 72.21&78.68\\
Deep3DBox \cite{mousavian20173d} &- &97.50 &\textbf{92.90}& - & 96.30 &88.75& -  & 80.40&76.76\\
3D-RCNN \cite{kundu20183d} &90.70 &\textbf{97.70 }&89.98& 89.10 & 96.50&\textbf{89.25} & \textbf{79.50}  & \textbf{80.70}&\textbf{80.07}\\
\hline
Our Method &\textbf{97.28} & {97.57}&92.58 &\textbf{93.70} & \textbf{96.70}&88.72 &{79.25}  &{80.45}& 76.85 \\
\hline
\end{tabular}
\end{center}
\label{tab:orientation}
\end{table*}

\begin{table*}[tb]
\caption{Comparisons of the 2D AP with the state-of-the-art methods on the KITTI Birds Eyed View validation dataset.}
\begin{center}
\begin{tabular}{|c|c|c|c|c|c|c|c|c|c|c|c|c|}
\hline
\multirow{3}*{Method} &  \multicolumn{6}{c|}{IoU = 0.5}&  \multicolumn{6}{c|}{IoU = 0.7}\\
\cline{2-13}
~ & \multicolumn{2}{c|}{ Easy} & \multicolumn{2}{c|}{Moderate} & \multicolumn{2}{c|}{Hard}  &\multicolumn{2}{c|}{ Easy} & \multicolumn{2}{c|}{Moderate} & \multicolumn{2}{c|}{Hard}  \\
\cline{2-13}
~ &  t/v 1 & t/v 2&  t/v 1 & t/v 2&  t/v 1 & t/v 2&  t/v 1 & t/v 2&  t/v 1 & t/v 2&  t/v 1 & t/v 2 \\
\hline\hline
3DOP \cite{chen20153d} &55.04 & - & 41.25 & - & 34.55  & - &12.63 & - &9.49 & - & 7.59  & -\\
Mono3D \cite{chen2016monocular} &30.50 & - & 22.39 & - & 19.16  & - &5.22 & - & 5.19 & - & 4.13  & -\\
Deep3DBox \cite{mousavian20173d} &- &30.02 & - &23.77 & -  & 18.83 &- & 9.99 & - &7.71 & -  &5.30\\
\hline
Our Method &32.57 & 33.37 &24.60 & 26.29 & 21.25  &21.57 & 9.50 & 10.45 & 8.02& 8.59 & 7.71  &7.43 \\
\hline
\end{tabular}
\end{center}
\label{tab:locationbird}
\end{table*}

\begin{table}[tb]
\caption{Comparisons of the Average Error of dimension estimation with state-of-the-art methods on the KITTI validation dataset.}
\vspace{-0.25cm}
\begin{center}
\begin{tabular}{|c|c|c|}
\hline
Method&  train/val 1 & train/val 2 \\
\hline\hline
3DOP \cite{chen20153d} &0.3527 & -\\
Mono3D \cite{chen2016monocular} &0.4251 & - \\
Deep3DBox \cite{mousavian20173d} &- &0.1934\\
\hline
Our Method &\textbf{0.1698} & \textbf{0.1465}\\
\hline
\end{tabular}
\end{center}
\label{tab:dimension}
\end{table}

For FQNet, we used the ImageNet pre-trained VGG-M model with $107\times 107$ input size to initialize the weights of convolutional layers. The projection was drawn using green color and linewidth of 1. For the classification pre-training process, we sampled 256 positive samples and 256 negative samples to train the network for each iteration with learning rates $10^{-4}$ for convolutional layers and $10^{-3}$ for fully connected layers. For the 3D IoU regression, we mapped the labels from $[0,1]$ to $[-1,1]$ for data balance.
We also added a random contour context information to the training image patch to increase the robustness of the model.

\subsection{Effectiveness}
To demonstrate that our proposed pipeline is not sensitive to the 2D detection results, we performed jitter to the MS-CNN 2D detection results and got a set of 2D bounding boxes with different Average Precision (AP). We compared the 3D detection performance (3D AP) of our method with the tight-constraint-based baseline, and the results are shown in Figure \ref{fig:sensitive}. We can see that the tight-constraint-based method is much more sensitive to the 2D detection AP, while our approach is much more robust.

We also studied our FQNet module by evaluating the 3D IoU regression performance. After dividing all proposals into ten levels based on the 3D IoU, we computed the average regression error for each level and drew the histogram in Figure \ref{fig:iou}. We can see that for the samples whose 3D IoU is around 0.4 to 0.5, the average estimation error is the lowest, which is about 0.05. Therefore, we can be sure that our FQNet have the ability to evaluate candidates.

\begin{table*}[tb]
\caption{Comparisons of the 3D AP with the state-of-the-art methods on the KITTI 3D Object validation dataset.}
\begin{center}
\begin{tabular}{|c|c|c|c|c|c|c|c|c|c|c|c|c|}
\hline
\multirow{3}*{Method} &  \multicolumn{6}{c|}{IoU = 0.5}&  \multicolumn{6}{c|}{IoU = 0.7}\\
\cline{2-13}
~ & \multicolumn{2}{c|}{ Easy} & \multicolumn{2}{c|}{Moderate} & \multicolumn{2}{c|}{Hard}  &\multicolumn{2}{c|}{ Easy} & \multicolumn{2}{c|}{Moderate} & \multicolumn{2}{c|}{Hard}  \\
\cline{2-13}
~ &  t/v 1 & t/v 2&  t/v 1 & t/v 2&  t/v 1 & t/v 2&  t/v 1 & t/v 2&  t/v 1 & t/v 2&  t/v 1 & t/v 2 \\
\hline\hline
3DOP \cite{chen20153d} &46.04 & - & 34.63 & - & 30.09  & - &6.55 & - &5.07 & - & 4.10  & -\\
Mono3D \cite{chen2016monocular} &25.19 & - & 18.20 & - & 15.52  & - &2.53 & - & 2.31 & - & 2.31  & -\\
Deep3DBox \cite{mousavian20173d} &- &27.04 & - & 20.55 & -  & 15.88 &- & 5.85 & - & 4.10 & -  & 3.84\\
\hline
Our Method &28.16&28.98 &21.02  & 20.71 &  19.91  &18.59 & 5.98 & 5.45 & 5.50 & 5.11 & 4.75  &4.45\\
\hline
\end{tabular}
\end{center}
\label{tab:location}
\end{table*}

\begin{figure*}[tb]
    \begin{center}
       \includegraphics[width=1.0\linewidth]{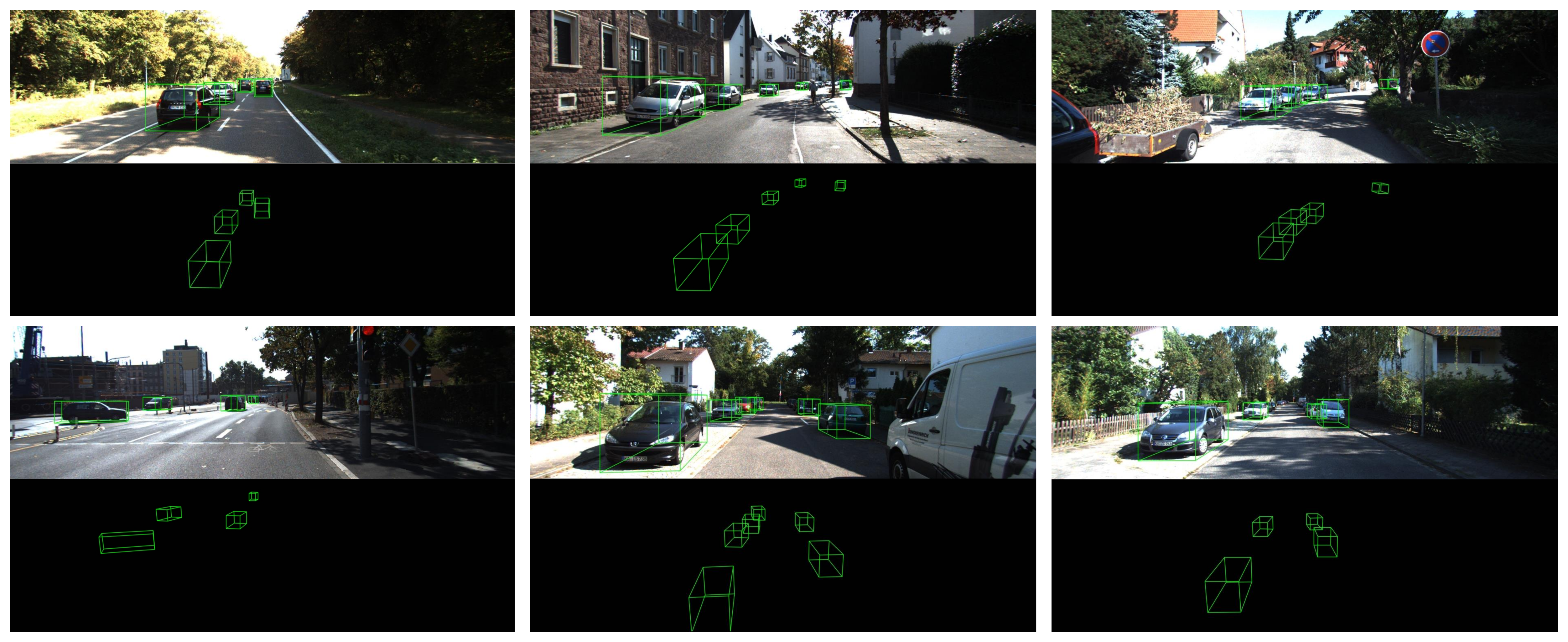}
    \end{center}
       \caption{The visualization result of our monocular 3D object detection method. We draw detection results in both 2D image and 3D space. }
    \label{fig:qualitative}
\end{figure*}

\subsection{Comparison with State-of-the-Arts}
We compared our proposed method with 6 recently proposed state-of-the-art 3D object detection methods on the KITTI benchmark, including 3DOP \cite{chen20153d}, Mono3D \cite{chen2016monocular}, 3DVP \cite{xiang2015data}, SubCNN \cite{xiang2017subcategory}, Deep3DBox \cite{mousavian20173d} and 3D-RCNN \cite{kundu20183d}. For fair comparisons, we used the detection results reported by the authors. All experiments were conducted on both two validation splits (different models are trained with the corresponding training sets).
\newline\newline
\noindent
\textbf{Orientation and Dimension Evaluation: }For orientation evaluation, we used the official metric of the KITTI dataset, which is the Average Orientation Similarity (AOS).
Our results are summarized in Table \ref{tab:orientation}. From the results, we see that our method achieves state-of-the-art method on both train/val 1 and train/val 2 experimental settings. And we can see that especially for the train/val 1 on Easy and Moderate setting, our method has a significant improvement comparing with existing methods.

For dimension evaluation, we use the average error defined as:
\begin{equation}
E_a = \frac{1}{N}\sum^N_{i=1} \sqrt{(\Delta w_i^2 +  \Delta h_i^2 + \Delta l_i^2)}
\end{equation}
Since the detection results and the ground truth are not one-to-one equivalents, we have to find the corresponding object in the ground truth which is closest to the detection result for computing $E_a$. Not all methods provided their experimental results, so we only compare our method with 3DOP \cite{chen20153d}, Mono3D \cite{chen2016monocular}, and Deep3DBox \cite{mousavian20173d}.
Our results are summarized in Table \ref{tab:dimension}. We can see that our methods have the lowest estimation error with an average dimension estimation error of about 0.15 meters, which demonstrate the effectiveness of our anchor based regression module.
\newline\newline
\noindent
\textbf{Location Evaluation:  }
For location evaluation, we first reported our results on the official evaluation metrics from KITTI Bird’s Eye View Evaluation, where AP for the bird’s eye view boxes is evaluated, which are obtained by projecting the 3D boxes to the ground plane and neglect the location precision on the Y-axis. From Table \ref{tab:locationbird}, we can see that our method outperformed Mono3D \cite{chen2016monocular} and Deep3DBox \cite{mousavian20173d} by a significant margin of about 3\%  improvement. Since 3DOP \cite{chen20153d} is a stereo-based method that can obtain depth information directly, so its performance is much better than pure monocular based methods.

We also conducted experiments on 3D Object Detection Evaluation, where the 3D AP metric is used to evaluate the full 3D bounding boxes. From Table \ref{tab:location}, we can see that our method ranks first among pure monocular based methods, and we even outperformed stereo-based 3DOP when 3D IoU threshold is set to 0.7.

\subsection{Qualitative Results}
Apart from drawing the 3D detection boxes on 2D images, we also projected the 3D detection boxes in the 3D space for better visualization. As shown in Figure \ref{fig:qualitative}, our approach can fit the object well and achieve high-precision 3D perception in various scenes with only one monocular image as input.

\section{Conclusions}
In this paper, we have proposed a unified pipeline for monocular 3D object detection. By using an anchor-based regression method, we achieved a high-precision dimension and orientation estimation. Then we perform dense sampling in the 3D space and project these samples on a 2D image. Through measuring the relation between the projections and object, our FQNet successfully estimates the 3D IoU and filters the suitable candidate. Both quantitative and qualitative results have demonstrated that our proposed method outperforms the state-of-the-art monocular 3D object detection methods. How to extend our monocular 3D object detection method for monocular 3D object tracking seems to be interesting future work.

\section*{Acknowledgement}
This work was supported in part by the National Natural Science Foundation of China under Grant 61822603, Grant U1813218, Grant U1713214, Grant 61672306, and Grant 61572271.
{\small
\bibliographystyle{ieee}
\bibliography{egbib}
}

\end{document}


\title{Deep Fitting Degree Scoring Network for Monocular 3D Object Detection \\ Supplementary Material}

\author{Lijie Liu$^{1,2,3,4}$, Jiwen Lu$^{1,2,3,*}$, Chunjing Xu$^{4}$, Qi Tian$^{4}$, Jie Zhou$^{1,2,3}$ \\
$^{1}$Department of Automation, Tsinghua University, China\\
$^{2}$State Key Lab of Intelligent Technologies and Systems, China\\
$^{3}$Beijing National Research Center for Information Science and Technology, China\\
$^{4}$Noah's Ark Lab, Huawei\\
{\tt\small liulj17@mails.tsinghua.edu.cn\qquad  \{lujiwen,jzhou\}@tsinghua.edu.cn,}\\
{\tt\small \{xuchunjing,tian.qi1\}@huawei.com}}

\maketitle
\thispagestyle{empty}
\newcommand\blfootnote[1]{%
\begingroup
\renewcommand\thefootnote{}\footnote{#1}%
\addtocounter{footnote}{-1}%
\endgroup
}
\blfootnote{$^{*}$   Corresponding Author}
\appendix
\section{Global Orientation}
\begin{figure}[tb]
    \begin{center}
       \includegraphics[width=1.0\linewidth]{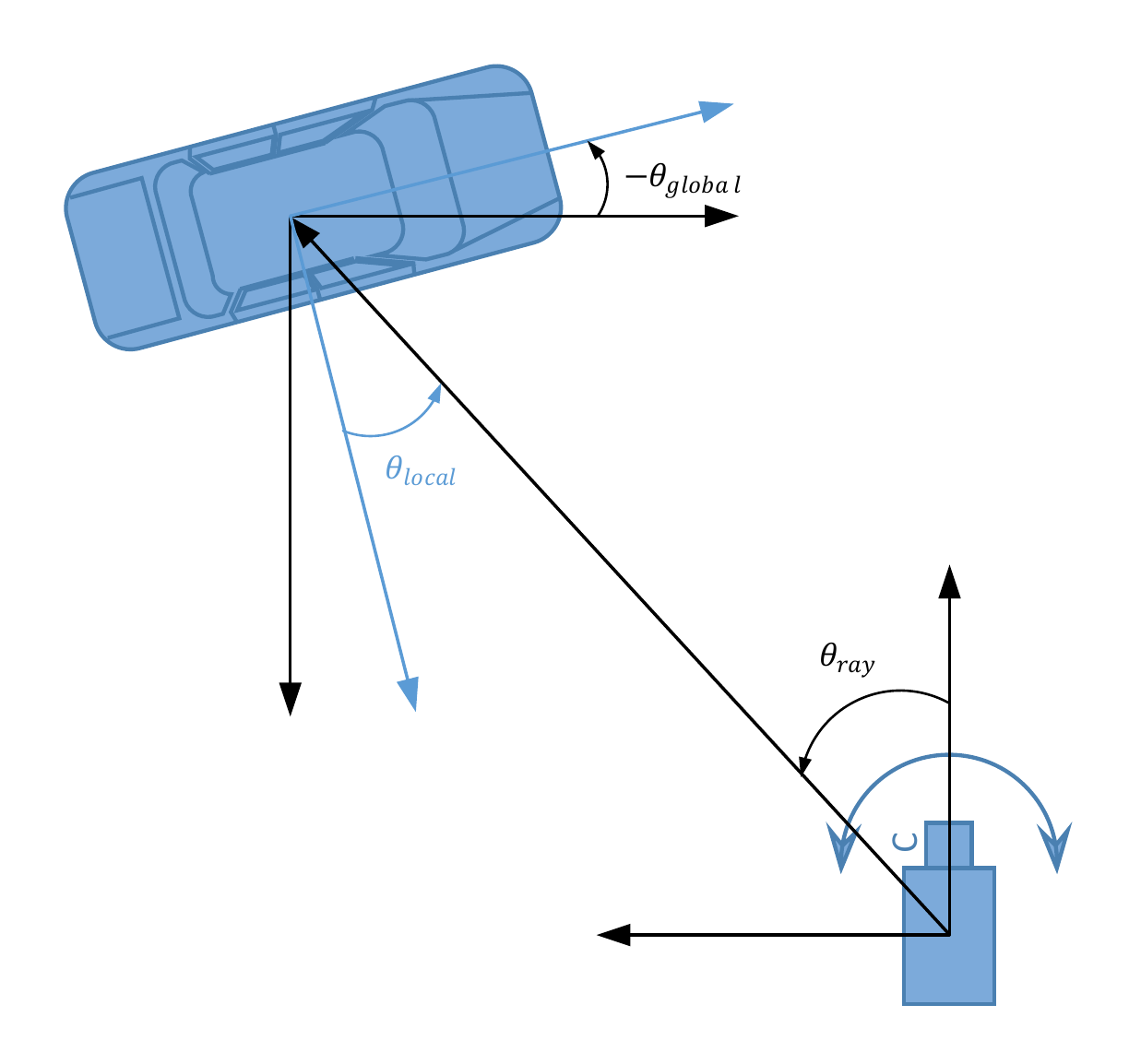}
    \end{center}
       \caption{Illustration of different orientations in bird view. For $\theta_{global}$, if the rotation is counterclockwise, it will have a negative measure. For $\theta_{local}$, its value is $0$ degree when the camera captures the right side of the object. The positive directions of $\theta_{local}$ and $\theta_{ray}$ are marked with arrows in the image.}
    \label{fig:supp1}
\end{figure}

Here we elaborate on more details about how to use local orientation to compute global orientation. In Section 3.1, we have explained how to regress the local orientation $\theta_{local}$, based on the appearance of the object. However, in Section 3.2, to calculate the rotation matrix $\bm{R}$, the global orientation $\theta_{global}$ is required. Denote the ray direction of the object as $\theta_{ray}$, as shown in Figure~\ref{fig:supp1}, we have:
\begin{equation}
    \theta_{ray} =  \theta_{local} + ( - \theta_{global})
\label{eqa:ld}
\end{equation}

Therefore, we only need to calculate $\theta_{ray}$ first, which is approximately proportional to the distance between the object center and the image center:
\begin{equation}
    \theta_{ray} \approx  k ( \frac{width}{2} - \frac{x_2+x_1}{2})
\label{eqa:ld}
\end{equation}
where $width$ is the width of the 2D image, $x_1$ and $x_2$ are the left and right boundaries of the 2D bounding box of the object (a more precise object center should be the center of the eight projected corners, which is $\frac{\sum_{i=1}^8 x_i}{8}$ ), and $k$ is the proportionality coefficient (Strictly speaking, the relationship between $\theta_{ray}$ and the distance between object center and image center is the $arctan()$ function, and we used proportional function here for a coarse estimate).

We regressed the value of $k$ using the training data by minimizing the following objective:
\begin{equation}
    k^\star = \arg\min\limits_{k} \sum_{i=1}^N( y_i - k ( \frac{width}{2} - \frac{x_{i2}+x_{i1}}{2}))^2
\label{eqa:ld}
\end{equation}
where $N$ is the number of training samples.

\begin{figure}[tb]
    \begin{center}
       \includegraphics[width=1.0\linewidth]{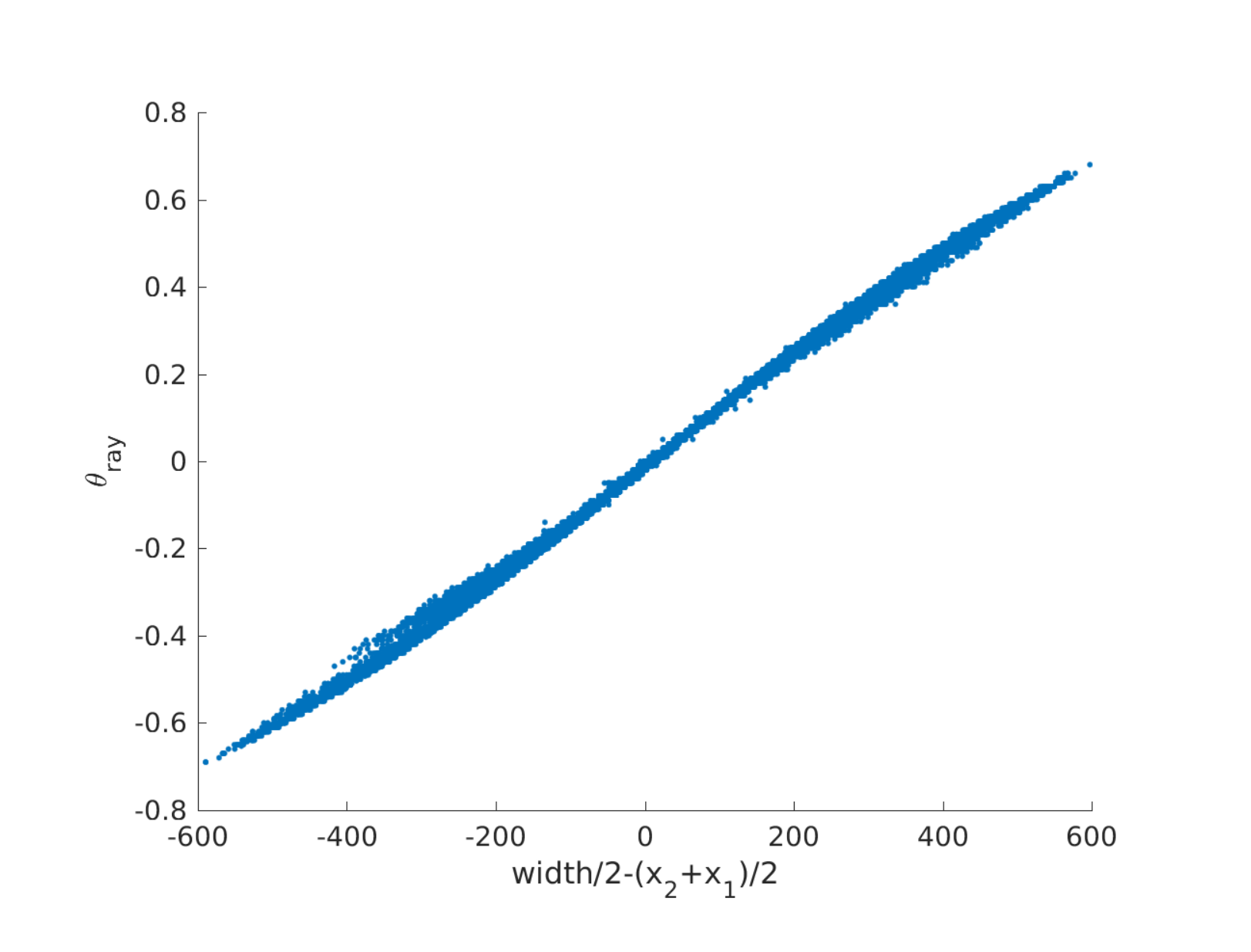}
    \end{center}
       \caption{Visualization of the relation between $\theta_{ray}$ and the distance to the center of the image.}
    \label{fig:supp2}
\end{figure}

We plotted the data points in Figure~\ref{fig:supp2} for better visualization, and the regression result is $k^\star = 0.0012408$.

\section{Tight Constraint}
\begin{figure}[tb]
    \begin{center}
       \includegraphics[width=1.0\linewidth]{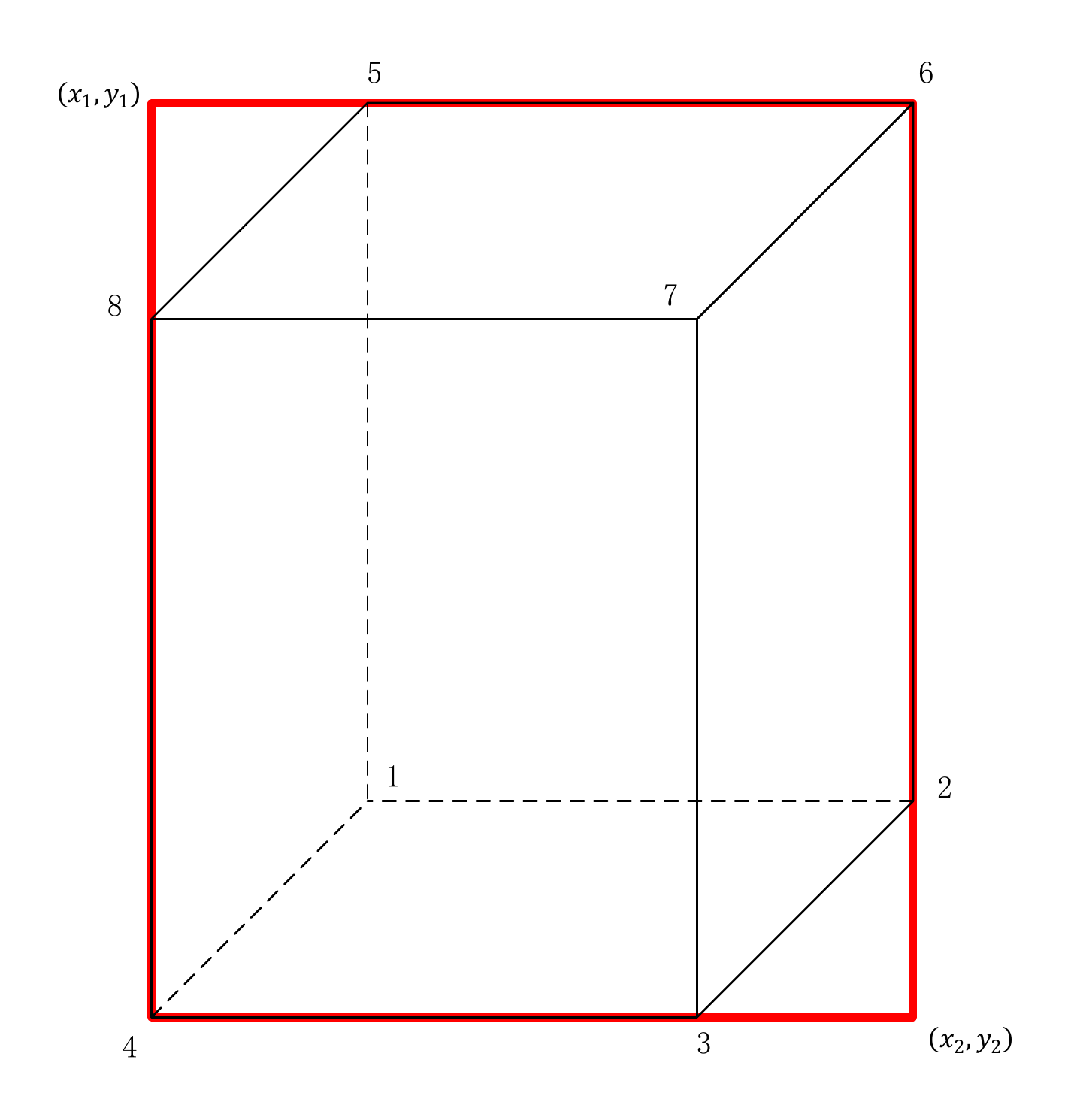}
    \end{center}
       \caption{Visualization of the tight constraint.}
    \label{fig:tight}
\end{figure}

As mentioned in Section 3.2, we used tight constraint similar in ~\cite{mousavian20173d, kundu20183d} to obtain the location of seed candidate. Here we provide more details about how to solve the location. The basic idea is to fit the projected 3D bounding box in the 2D bounding box tightly, which means the circumscribed rectangle of the projected 3D bounding box should be as close to the 2D bounding box as possible (Figure~\ref{fig:tight}).

Using regression module, we have estimated the dimension and orientation of the object, which means the 3D model is fixed and only the location is unknown. Denote the location as $\bm{T} = [t_x, t_y, t_z]^T$, we have:
\begin{equation}
    \bm{K} \begin{bmatrix}
 \bm{R} & \bm{T} \\
 \bm{0}^T & 1
 \end{bmatrix} \bm{X}_{3d}  = \bm{X}_{2d}
\label{equ:1}
\end{equation}
and Equation (\ref{equ:1}) can be rewritten into the form of $\bm{A}\bm{x} = \bm{b}$:
\begin{equation}
    \bm{K} \begin{bmatrix}
 \bm{I} & \bm{R}\bm{X}_{3d} \\
 \bm{0}^T & 1
 \end{bmatrix} \begin{bmatrix}
t_x \\
t_y\\
t_z\\
1
 \end{bmatrix}  = \bm{X}_{2d}
\end{equation}

For each boundary of the 2D bounding box ($x_1,y_1,x_2,y_2$), it has to be touched by one projected vertices, so there are $8^4 = 4096$ settings in total \footnote{This number can be reduced to 64 by using some prior knowledge, such as normally cars won't turn upside down, so vertices 1 won't touch $y_1$.}. For each setting, we can have 4 equations, but we only have 3 unknown, so it is an overdetermined equation which can be solved by least square method. We computed the residual for each setting, discarded the settings in which the projection of the 3D bounding box exceeded the 2D bouding box, and chose the setting with the least error as the solution.

\section{Tradeoffs of the System}
\begin{figure}[htb]
    \begin{center}
       \includegraphics[width=1.0\linewidth]{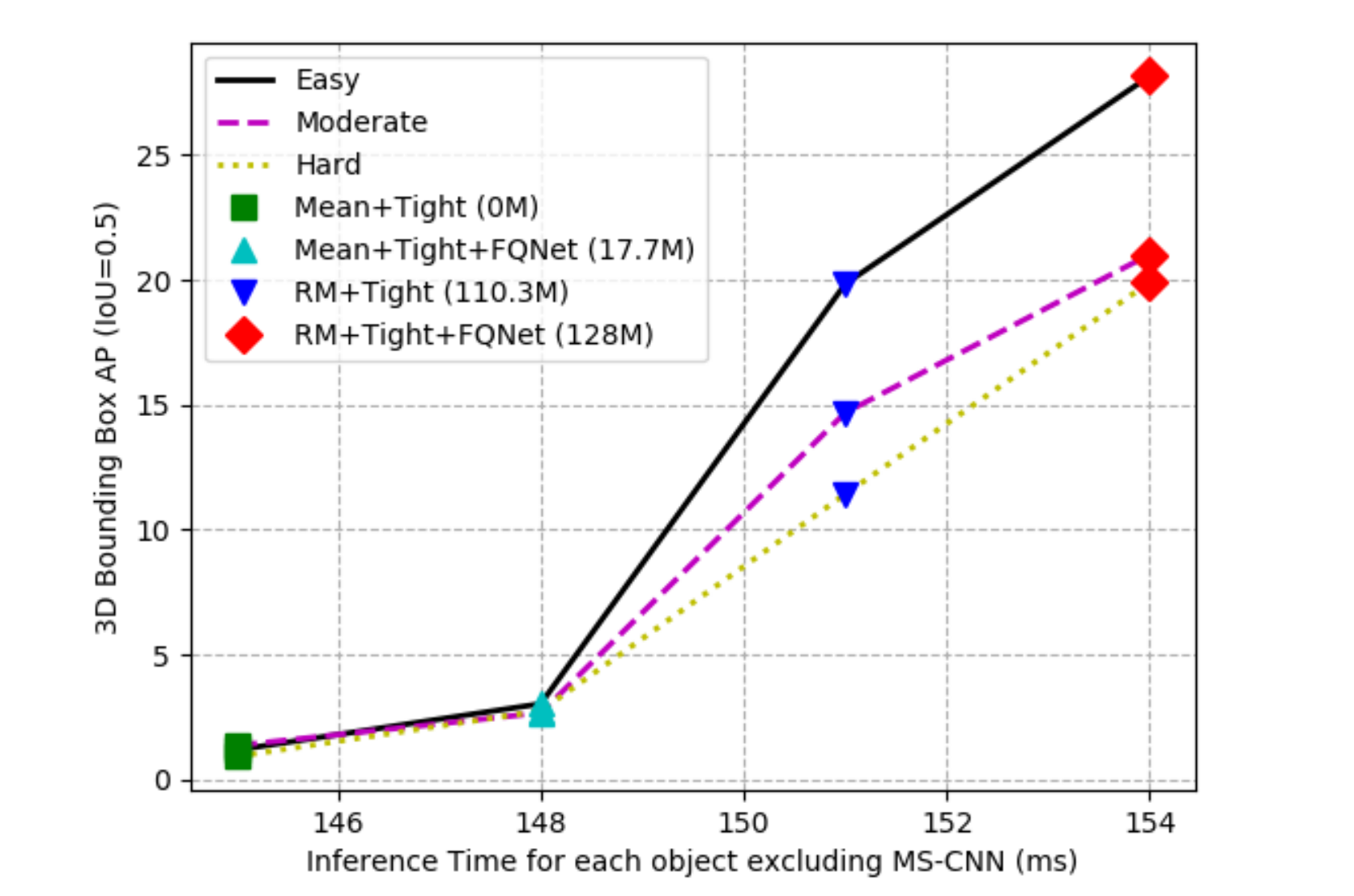}
    \end{center}
\vspace {-0.4cm}
       \caption{Speed versus accuracy on the KITTI validation set.}
    \label{fig:sensitive}
\end{figure}

Figure \ref{fig:sensitive} shows the accuracy against latency for different model configurations. In Mean+Tight, mean dimension and mode of the orientation\footnote{Most cars in the KITTI dataset are directed towards the street.} are used as estimation, and location is computed using tight constraint. Mean+Tight+FQNet applies our FQNet to refine the Mean+Tight detection results. RM+Tight refers to our regression module.  RM+Tight+FQNet is the final version of our proposed method. Our method achieved about 30\% improvement compared to Mean+Tight baseline, with a computation burden of 0.009s. We found that solving the tight constraints (0.145s) cost much more time than the forward process of RM and FQNet, so exploring a faster way to locate the seed candidate can speed up reasoning process dramatically. All experiments were conducted based on the MS-CNN [5] 2D detection boxes which costs 0.4s. We also reported the model size of our proposed RM and FQNet.

\vspace {-0.37cm}
\vspace {-0.29cm}

\section{Contributions of Each Module}
The proposed method can be divided into two modules: regression module (RM) and FQNet. As shown in Figure \ref{fig:sensitive}, two modules worked together to achieve the final results. We found that 2D regression accuracy had a great effect on the location estimation process. When using Mean+Tight as seed candidate, FQNet only gained 2\% improvement. While with RM+Tight, FQNet can achieve 8\% improvement. Another experiment also demonstrated that 2D regression accuracy is of great importance: when we replaced RM with ground truth dimension and orientation, GT+Tight can achieve 54.47\% in 3D AP. That is why we studied regression module to increase regression accuracy as much as possible.

\section{More Implementation Details}
For the 2D detection method, we used the code provided by MS-CNN \cite{cai2016unified} to get 2D bounding boxes, and filtered the results through abandoning the detections whose score is lower than 0.1.

Our experiments were conducted on the following specifications: i7-4790K CPU, 32GB RAM, and NVIDIA GTX1080Ti GPU using Python and PyTorch toolbox. In our settings, the whole pipeline runs at about 0.3\footnote{since for different testing image, there are different numbers of objects, so our running speed may fluctuate.} frames per second .

\section{Video Demo}
We refer the reader to the attached video for more visualizations of our results. We note that to create the video no temporal
information is used, and all results are obtained using a single monocular image. The sequences are from the KITTI object tracking benchmark, with index of 0001, 0004, and 0007.

{\small
\bibliographystyle{ieee}
\bibliography{egbib}
}